\PassOptionsToPackage{table}{xcolor}
\documentclass[11pt, copyright, logo, secondlogo]{template/lark}
\usepackage[authoryear, sort&compress, round]{natbib}
\usepackage{bbm}
\usepackage{enumitem}
\usepackage{xspace}
\usepackage{graphicx}
\usepackage{tabularx}
\usepackage{booktabs}
\usepackage{subcaption}
\usepackage{xcolor}
\usepackage{hyperref}
\usepackage{verbatim}
\usepackage{float}
\usepackage{wrapfig}
\usepackage{capt-of}
\usepackage{makecell}
\usepackage{multirow}
\usepackage{url}
\usepackage{algorithm}
\usepackage{algpseudocode}
\usepackage{cleveref}

\usepackage{amsmath,amsfonts,bm}

\def\1{\bm{1}}

\DeclareMathAlphabet{\mathsfit}{\encodingdefault}{\sfdefault}{m}{sl}
\SetMathAlphabet{\mathsfit}{bold}{\encodingdefault}{\sfdefault}{bx}{n}

\definecolor{linkblue}{RGB}{0,92,175}
\hypersetup{
  hypertexnames=false,
  colorlinks=true,
  linkcolor=linkblue,
  citecolor=linkblue,
  urlcolor=linkblue
}
\definecolor{tablegroupblue}{RGB}{239,245,249}

\newtheorem{proposition}{Proposition}
\newtheorem{assumption}{Assumption}
\crefname{assumption}{Assumption}{Assumptions}

\newcommand{\SpecAgent}{\textsc{SpecAgent}\xspace}

\title{Trajectory-Level Speculation with Semantic Verification for Tool-Using Agents}

\author[1,2]{Zilin Wang}
\author[2,3,\dag,\ddag]{Yinhong Liu}
\author[1,2]{Chao Chen}
\author[2]{Qun Liu}
\author[1,4,\dag]{Zhijiang Guo}

\affil[1]{LARK, HKUST (GZ)}
\affil[2]{Yinwang}
\affil[3]{University of Cambridge}
\affil[4]{HKUST}

\authornotes{$^{\dag}$Corresponding authors. $^{\ddag}$Project leader.}
\correspondingauthor{Yinhong Liu (yl535@cam.ac.uk), Zhijiang Guo (zhijiangguo@hkust-gz.edu.cn)}
\githubpage{https://github.com/LARK-AI-Lab/SpecAgent}

\begin{abstract}
Large language model agents solve tasks through long sequences of tool calls,
and the prevailing design has a single strong model generate every step,
although many steps are routine enough for a much smaller model. Existing
ways of sharing this work keep the strong model as the generator: routers
and collaborative agents decide which model handles each query or step, and speculative methods check a small
model's drafts one step at a time, usually against the strong model's own
action. We
introduce \SpecAgent{}, a collaboration protocol that turns the strong model
from the author of each step into a reviewer of steps drafted by a weak
model. The weak model executes read-only tool calls as it drafts, so that
every step is grounded in real observations, and holds the first action that
would modify the environment. The strong model then verifies the drafted
segment in a single prefill pass, commits the longest prefix it judges
semantically valid, and repairs the first rejected step. Because verification
replaces autoregressive decoding with parallel prefill, the strong model
decodes only short verdicts and occasional repairs rather than every step.
On $\tau^2$-Bench, \SpecAgent{} comes within one point of the strong model's
success rate while roughly halving memory traffic and wall time, and on
AutomationBench and VitaBench it recovers 44--75\% of the quality gap while
lowering both. In an edge--cloud deployment on MCP-Atlas with
three cloud models, it closes 53--75\% of the weak--strong quality gap while
the strong model generates at most half as many tokens. These results suggest
that a strong model need not author every step of an agent: reviewing
grounded drafts retains most of its quality while cutting both latency and
memory traffic.
\end{abstract}
 
\begin{document}
\maketitle

\section{Introduction}
\label{sec:introduction}

Language-model agents now solve tasks through long sequences of tool calls and
user turns \citep{deng2026clarification}, from tens of turns in customer service to up to a hundred in
everyday life services (Table~\ref{tab:interaction_lengths}). The prevailing
design has one strong model generate every step, so serving cost grows with
the length of the trajectory and is dominated by the strong model's
token-by-token decoding. Yet many steps are routine, such as looking up an
order, reading a record, or searching again, and a much smaller model could
often produce them.

\begingroup
\setlength{\intextsep}{6pt}
\begin{table}[H]
  \centering
  \caption{Representative trajectory length per task for frontier agents, in
  turns (messages from the agent, the user, and tools).}
  \label{tab:interaction_lengths}
  \defcitealias{barres2025tau2}{$\tau^2$-Bench}
  \defcitealias{bandi2026mcpatlas}{MCP-Atlas}
  \defcitealias{shepard2026automationbench}{AutomationBench}
  \defcitealias{he2025vitabench}{VitaBench}
  \small
  \setlength{\tabcolsep}{3pt}
  \renewcommand{\arraystretch}{1.12}
  \begin{tabular*}{\textwidth}{@{\extracolsep{\fill}}cccc@{}}
    \toprule
    \textbf{Real-World MCP Services} &
    \textbf{Transactional Interaction} &
    \textbf{Business Workflows} &
    \textbf{Life Services} \\
    \citepalias{bandi2026mcpatlas} &
    \citepalias{barres2025tau2} &
    \citepalias{shepard2026automationbench} &
    \citepalias{he2025vitabench} \\
    \midrule
    20--30 turns & 30--80 turns & 40--60 turns & 50--100 turns \\
    \bottomrule
  \end{tabular*}
\end{table}
\endgroup

Existing ways of sharing this work keep the strong model as the generator.
Routers pick one model per query
\citep{chen2023frugalgpt,aggarwal2024automix,ding2024hybridllm,ong2024routellm},
and collaborative agents decide, step by step, which model should act
\citep{gao2026agentcollab}, so every step is still written by one model from scratch. Speculative agents
carry the draft-and-verify idea of speculative decoding
\citep{leviathan2023speculative} over to actions
\citep{hua2024isp,ye2026specactions}, but usually keep a draft only when it matches the
strong model's own action, so the strong model must still generate each
reference step and its work is overlapped rather than removed. Recent variants
replace matching with a step-level critic
\citep{huang2025spagent,zhong2026dualspec}, but they check one predicted step
at a time in search-style tasks whose actions do not change the environment. Exact matching is also
too strict for agents: different queries, lookup orders, or argument choices
can all lead to a valid solution \citep{liu2026schemabias}, and progress that differs from the strong
model's choice is discarded. We therefore ask: \emph{can a
strong model review several steps drafted by a weak model, and keep what is
valid, instead of generating every step itself?}

\begin{figure}[t]
  \centering
  \includegraphics[width=\linewidth]{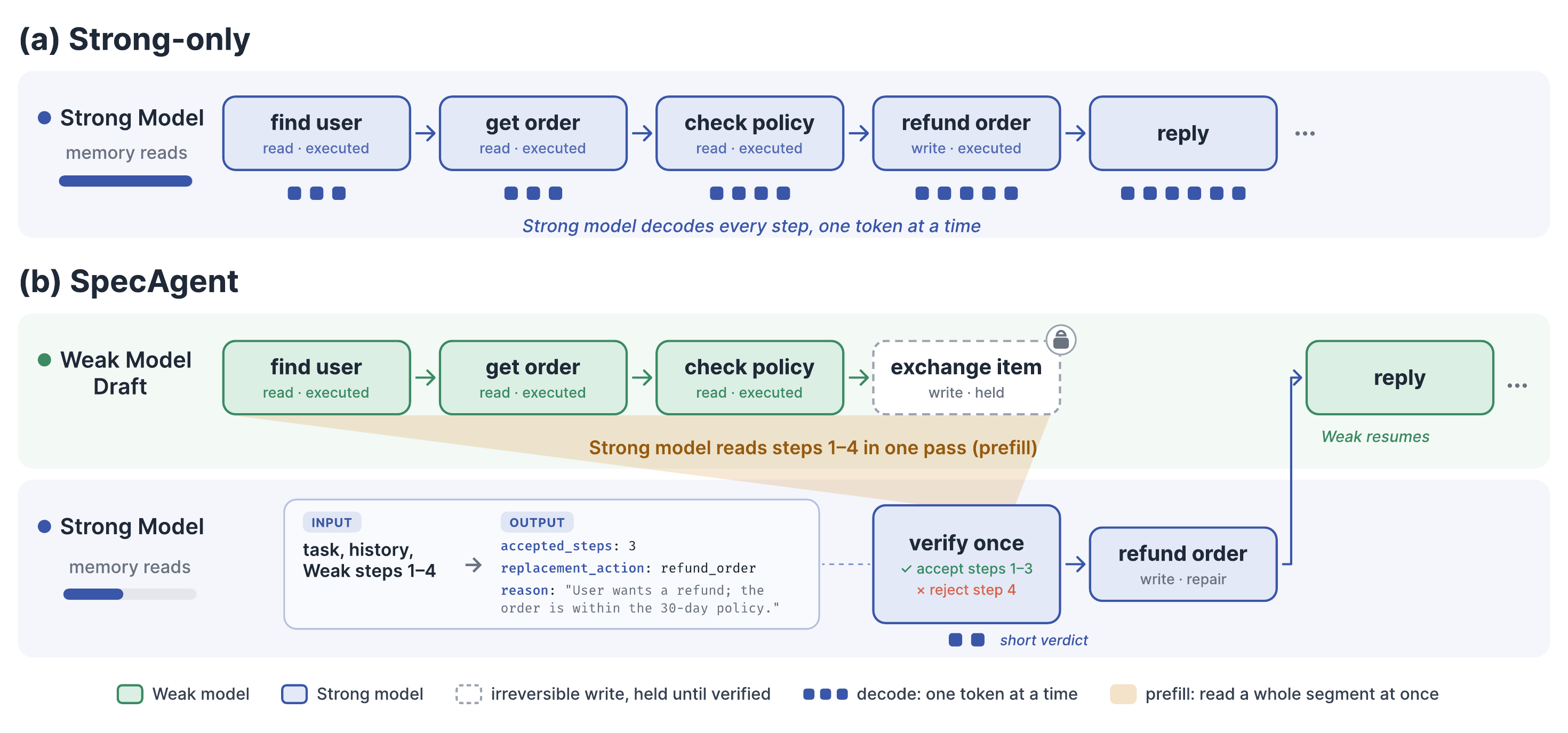}
  \caption{(a) A Strong-only agent generates every step by decoding.
  (b) \SpecAgent{} lets the Weak model draft ahead, running reads at once and
  holding the first write. The Strong model reviews the draft in a single
  prefill pass, keeps the valid steps, and replaces the rejected write; the
  Weak model then resumes. Bars indicate relative Strong-model memory reads.}
  \label{fig:specagent_overview}
\end{figure}

We answer it with \SpecAgent{}, in which a small \emph{Weak} model drafts the
trajectory and a large \emph{Strong} model reviews it, separating acting from
committing (Figure~\ref{fig:specagent_overview}). Three properties of agents shape its
design. First, a draft is meaningful only if it is grounded, because later
steps depend on what earlier tool calls return; the Weak model therefore
executes read-only calls as it drafts. Second, some actions cannot be undone,
so the Weak model holds the first state-changing action, such as changing an
order, and the Strong model reviews the segment before
it runs, commits the longest prefix it judges semantically valid, and repairs
the first rejected step. Third, review must cost less than generation even
though the reviewer reads the full context: the Strong model reads a segment
in one parallel prefill pass, whereas generating it would require decoding
every token.

We compare against \emph{Strong-only} and \emph{Weak-only}, the single-model
agents in which one model generates every step, and evaluate complete
quality--resource operating points, counting both models and keeping compute,
memory traffic, and wall time as separate axes. On
$\tau^2$-Bench, \SpecAgent{} comes within one point of Strong-only success
while reducing estimated memory traffic to 43--57\% and wall time to
41--54\% of Strong-only; on AutomationBench and VitaBench it recovers 44\% and 75\% of the
Weak--Strong quality gap, again with lower memory traffic and wall time. In an
edge--cloud deployment on MCP-Atlas with three cloud Strong models, it closes
53--75\% of the gap while the Strong model generates at most half as many
tokens. The saving comes from decoding rather than arithmetic: estimated FLOPs
rise by 30--38\% on $\tau^2$ and AutomationBench and by 87\% on VitaBench, so \SpecAgent{} suits memory-bandwidth-bound serving.

Our contributions are as follows.
\begin{itemize}[leftmargin=*, itemsep=0pt, topsep=2pt, parsep=0pt]
    \item \textbf{Framework.} \SpecAgent{}, a weak--strong protocol in which
    the Weak model drafts with real read-only tool calls and holds the first
    state-changing action, and the Strong model commits the longest prefix it
    judges semantically valid and repairs the first rejected step, together
    with a cost model that predicts when replacing Strong decoding by prefill
    saves memory traffic and time.
    \item \textbf{Evaluation.} Experiments on $\tau^2$-Bench, AutomationBench, VitaBench,
    and an edge--cloud deployment on MCP-Atlas with three cloud Strong models,
    comparing complete quality--resource operating points against Weak--Strong
    collaboration and speculative baselines.
    \item \textbf{Analysis.} An accounting of where the savings arise:
    measured Strong traffic and time follow the cost model, the saving comes
    from decoding, not computation, and it shrinks as acceptance falls.
\end{itemize}
\section{Related Work}
\label{sec:related_work}

\paragraph{Speculative execution.}
Speculative decoding lets a small model draft tokens that a large model
verifies in parallel while preserving its output distribution
\citep{leviathan2023speculative,chen2023accelerating}; it is complementary to
\SpecAgent{}, accelerating each model's own decoding. The draft-then-verify
pattern has since been applied to reasoning and search
\citep{wang2025speccot,wang2025specsearch,fu2025lookahead} and to agents, where
a fast model predicts plans, actions, or tool calls
\citep{hua2024isp,guan2025dsp,ye2026specactions,nichols2025spectool,hooper2026specinteraction,ji2026specreason},
sometimes with relaxed or semantic matching
\citep{huang2025spagent,zhong2026dualspec}. These methods target latency and
keep a draft when it agrees with the large model's own step. \SpecAgent{} instead verifies multi-step, grounded segments in one pass and
checks only before actions that change the environment.

\paragraph{Routing and weak--strong collaboration.}
Cascades and routers pick one model per query, including edge--cloud
deployments \citep{chen2023frugalgpt,aggarwal2024automix,ding2024hybridllm,ong2024routellm};
agent-level methods hand control between models during a trajectory
\citep{gao2026agentcollab,hemadri2026r2v,anon2026steplevelcua}. In both, the
model in control generates each step; in \SpecAgent{}, the Strong model
decides how much drafted progress to keep and generates only repairs.

\paragraph{Verification versus generation.}
Checking a candidate is often easier than producing one, as verifiers and
process reward models show for reasoning
\citep{cobbe2021training,lightman2024verify}. \SpecAgent{} applies this
asymmetry to agent trajectories, with the Strong model as the verifier, and
exploits its cost counterpart: prefill processes many tokens per weight read,
whereas decoding rereads them for every token
\citep{pope2023efficiently,kwon2023vllm,zheng2024sglang}.
\section{\SpecAgent{}}
\label{sec:method}

\SpecAgent{} distinguishes steps by whether their effects can be undone. A
read leaves the state untouched, so the Weak model can execute it immediately,
and a wrong read costs no more than being discarded; a state-changing action
is where an error becomes permanent, and it is the only point at which
\SpecAgent{} stops for the Strong model. This lets the Weak model speculate
over whole stretches of the trajectory while keeping execution safe. When the
Strong model steps in, it judges whether the drafted progress is valid rather
than whether it matches what it would have done, so it reviews the trajectory
instead of writing it (Figure~\ref{fig:specagent_overview}).

\subsection{Problem Formulation}
\label{sec:method_setting}

\begin{wrapfigure}[29]{r}{0.5\linewidth}
\vspace{-1.2em}
\begin{minipage}{\linewidth}
\begin{algorithm}[H]
\caption{\SpecAgent{}: draft, verify, repair}
\label{alg:specagent}
\footnotesize
\begin{algorithmic}[1]
\Require Task $u$, Weak policy $\pi_W$, Strong verifier $V_S$, Strong policy
$\pi_S$, chunk size $K$, adapter $E$
\State $H\gets(u)$
\While{the task is not complete}
  \State $\hat\tau\gets[\,]$
  \Loop
    \State $\hat a\gets\pi_W(H\oplus\hat\tau)$
    \If{$E.\textsc{StateChanging}(\hat a)$}
      \State Append $(\hat a,\bot)$ to $\hat\tau$; \textbf{break}
    \EndIf
    \State $\hat o\gets E.\textsc{Execute}(\hat a)$; append $(\hat a,\hat o)$ to
      $\hat\tau$
    \If{$|\hat\tau|=K$}
      \State $H\gets H\oplus\hat\tau$; $\hat\tau\gets[\,]$
    \EndIf
  \EndLoop
  \State $(j,r,\tilde a)\gets V_S(H,\hat\tau)$
  \If{$j=|\hat\tau|$}
    \State $H\gets H\oplus\operatorname{Prefix}_{j-1}(\hat\tau)$
    \State Execute the held action; append its result to $H$
  \Else
    \State $H\gets H\oplus\operatorname{Prefix}_j(\hat\tau)$
    \If{$E.\textsc{Valid}(\tilde a)$}
      \State $c\gets\tilde a$
    \Else
      \State $c\gets\pi_S(H,r)$
    \EndIf
    \State Execute $c$; append $(c,o_c)$ to $H$
  \EndIf
\EndWhile
\end{algorithmic}
\end{algorithm}
\end{minipage}
\vspace{-1.0em}
\end{wrapfigure}
 We model a tool-using task as a partially observable Markov decision process.
The environment has a hidden state $s$, such as a database or a workspace,
that the agent cannot see directly. At each step the agent issues an action
$a_t$, typically a tool call, and receives an observation $o_t$. A policy
therefore conditions on the task $u$ and the history
\begin{equation}
H_t=\bigl(u,(a_1,o_1),\ldots,(a_t,o_t)\bigr).
\label{eq:confirmed_history}
\end{equation}
Two policies share this interface: a Weak model $\pi_W$ that is cheap to run
and a Strong model $\pi_S$ whose quality we want to preserve.

Actions fall into two classes. A \emph{read}, such as a search or a lookup,
returns an observation and leaves the state unchanged. A
\emph{state-changing} action, such as modifying a record or submitting a
final result, alters the state. The distinction determines what can be
undone, and it is the principle behind \SpecAgent{}. An error that stays in
the history is recoverable: a wrong read is removed by deleting it from
the history, and because reads leave the task state unchanged, the environment is as it was
(Appendix~\ref{sec:appendix_adapters}). An error that reaches the environment is
not, since no edit to the history withdraws a changed record or a submitted
result. Speculation can therefore proceed freely while execution remains
reversible, and verification is needed exactly at the boundary where an
error would become irreversible: earlier checks spend the Strong model on
steps that could still be undone, and later ones come too late.

\subsection{Speculative Execution Protocol}
\label{sec:method_protocol}

\paragraph{Draft.}
From $H_t$, the Weak model drafts a chunk of at most $K$ steps,
\begin{equation}
\hat\tau=\bigl((\hat a_1,\hat o_1),\ldots,(\hat a_{n-1},\hat o_{n-1}),
(\hat a_n,\bot)\bigr).
\label{eq:candidate}
\end{equation}
Reads execute immediately, so each step is chosen after seeing real
observations rather than imagined ones. The chunk ends at the first
state-changing action $\hat a_n$, which is held rather than executed, since
every later step would depend on its outcome. A chunk that reaches $K$ steps
with only reads is committed without review, as the reversibility principle
allows.

\paragraph{Verify.}
The Strong model reads the history and the chunk in a single pass and
returns a verdict
\begin{equation}
(j,r,\tilde a)=V_S(H_t,\hat\tau), \qquad j\in\{0,\ldots,n\},
\label{eq:verification}
\end{equation}
where $j$ is the length of the longest acceptable prefix, $r$ explains the
first rejected step, and $\tilde a$ optionally replaces it. The verdict is a
prefix because each drafted step conditions on the observations of the steps
before it: once a step is wrong, later steps rest on a mistaken premise and
cannot be trusted even if they look valid. Acceptance is
\emph{semantic}: the verifier asks whether each step is a valid move toward
the task, not whether it is the step the Strong model would have written.
Tool-using tasks admit many valid trajectories, so semantic review retains
progress that exact matching would discard. If $j=n$, the whole chunk is
committed and the held action is executed.

\paragraph{Repair.}
A rejection should cost the Strong model as little generation as possible.
Many rejected steps are a single tool call with a wrong argument or target,
which the verifier can state while explaining the rejection; the verdict then
carries a replacement $\tilde a$, and the repair adds no Strong call. Some
repairs cannot be written that way, for instance when the fix spans several
calls or the proposed replacement fails the adapter's validity checks; only
then does the Strong model take one corrective turn from the history and the
reason $r$. The fused path keeps the common case cheap, and the fallback keeps
every rejection repairable. A repair is the Strong model's own step, so it is
committed without a second review. The rejected suffix, including the held
action, is discarded, so the next draft starts from a history of committed
steps only. Algorithm~\ref{alg:specagent} gives the full loop.

\subsection{Efficiency Analysis}
\label{sec:method_cost}

Autoregressive decoding generates one token per forward pass, and each pass
streams the model's active weights from GPU memory, so decoding is bound by
memory bandwidth rather than arithmetic~\citep{pope2023efficiently}. A
prefill pass processes many tokens against a single weight read. We make this
the basis of a simple cost model.

\begin{assumption}[Decode-bound serving]
\label{ass:decode_bound}
While a Strong replica decodes with a batch small enough to remain
memory-bound, the time and memory traffic of a Strong call are approximately
proportional to its number of weight passes: one per decoded token and one
for the prefill.
\end{assumption}

\begin{assumption}[Per-task acceptance]
\label{ass:acceptance}
As a first-order model, each task $u$ has an average per-step acceptance rate
$p_u$, related to the task's difficulty, the Weak model's competence, and the
Strong verifier's preferences; the steps of a reviewed chunk of length $n$ are
accepted independently.
\end{assumption}

Let $d$ be the weight passes of a Strong-only step, $v$ those of a verdict,
and $\rho$ the probability that a rejection needs a separate repair turn
because no valid replacement was fused into the verdict. The comparison is
intuitive. In Strong-only, every committed step costs a full decoded
step. In \SpecAgent{}, the Strong model instead decodes one short verdict per
reviewed chunk, shared by all the steps the chunk commits, and a repair only
when a step is rejected. Its cost per committed step should therefore shrink
as verdicts get shorter, as more of each chunk is accepted, and as fewer
rejections need a separate repair turn.
Proposition~\ref{prop:strong_decode} makes this precise.

\begin{proposition}[Strong decoding]
\label{prop:strong_decode}
Under \cref{ass:decode_bound,ass:acceptance}, on task $u$ the Strong model's
time and memory traffic per committed step, relative to Strong-only, are
\begin{equation}
R_u=\frac{1-p_u}{1-p_u^n}\cdot\frac{v}{d}+\rho\,(1-p_u).
\label{eq:strong_ratio}
\end{equation}
If task $u$ commits $T_u$ steps in both systems, the ratio over a task
distribution is the step-weighted mean
$R=\mathrm{E}_u[T_uR_u]/\mathrm{E}_u[T_u]$.
\end{proposition}

$R_u$ falls as $p_u$ rises, from $v/d+\rho$ at $p_u=0$ to $v/(nd)$ as
$p_u\to1$, so the saving concentrates on tasks where the Weak model's drafts
suit the verifier; read-only chunks committed without review only lower $R$
further. Saving Strong time does not by itself shorten
the task: every committed step must first be drafted, so the Weak model's
decoding replaces part of the Strong model's, and \SpecAgent{} is faster only
if a Weak step is cheap relative to a Strong one.

\begin{proposition}[Latency]
\label{prop:latency}
Under the same assumptions, let $\kappa$ be the time of a Weak drafting step
relative to a Strong-only step, and ignore tool latency. On task $u$ the
end-to-end time per committed step, relative to Strong-only, is
\begin{equation}
L_u=R_u+\kappa\,\frac{n(1-p_u)}{1-p_u^n},
\label{eq:latency_ratio}
\end{equation}
and the ratio over tasks is again the step-weighted mean of $L_u$.
\end{proposition}

The second term counts drafted steps per committed step and approaches one as
$p_u\to1$. Proofs are in Appendix~\ref{sec:appendix_proofs}, and
Section~\ref{sec:analysis} examines where the savings arise.
\section{Experiments}
\label{sec:evaluation}

We evaluate \SpecAgent{} on agent tasks that mix reads, state-changing
actions, and user-visible replies. The experiments ask three questions.
\textbf{Q1:} how much of the Weak--Strong quality gap does semantic prefix
verification recover? \textbf{Q2:} which resources does it save, and which
does it not? \textbf{Q3:} how does it compare with existing Weak--Strong
collaboration methods at their own operating points?

\subsection{Setup}
\label{sec:evaluation_shared}

\paragraph{Benchmarks.}
$\tau^2$-Bench Airline (50 tasks) and Retail (114 tasks) test multi-turn
customer service against a simulated user, mixing database reads, mutations,
and replies \citep{barres2025tau2}. We run three independent trajectories per
task and report the two domains separately because their action mix and
difficulty differ. AutomationBench tests long business workflows whose
success is checked on the final environment state
\citep{shepard2026automationbench}; we aggregate its 600 public tasks
into one block, with its development domain excluded, and run one trajectory
per task. VitaBench Delivery
(100 tasks) tests food-delivery service against a simulated user, judged by an
LLM against task rubrics \citep{he2025vitabench}; we run two trajectories per
task. MCP-Atlas, run against
real tool servers in an edge--cloud deployment, is reported separately
(Section~\ref{sec:evaluation_mcpatlas}).

\paragraph{Models and methods.}
On $\tau^2$, AutomationBench, and VitaBench, the Weak model is Qwen3.5-4B and the Strong
model is Qwen3.5-397B-A17B \citep{qwen2026qwen35}.
Weak-only and Strong-only run the same agent loop with a single model.
AgentCollab \citep{gao2026agentcollab}, the Weak--Strong collaboration
baseline, escalates to Strong using its published self-evaluation signal.
Speculative Actions \citep{ye2026specactions} keeps Strong as the actor and
speculates on upcoming steps to hide latency; we port the authors' released
agent to $\tau^2$ (Appendix~\ref{sec:appendix_specactions}). We do not port it to
AutomationBench or VitaBench. All methods run on the
same Weak and Strong deployments. \SpecAgent{} reviews at state-changing
actions with chunk size $K=6$ on $\tau^2$ and VitaBench and $K=8$ on AutomationBench; other
adapter settings and all prompts are listed in
Appendix~\ref{sec:appendix_adapters}.

\paragraph{Serving.}
On $\tau^2$, AutomationBench, and VitaBench, all models run on NVIDIA H20 GPUs: the Weak
model on a single GPU with SGLang \citep{zheng2024sglang}, and the Strong
model with 16-way tensor parallelism. Every method runs eight tasks
concurrently against the same deployments, so wall time is compared under a
common load. Decoding settings and the user simulators are given in
Appendix~\ref{sec:appendix_setup}.

\paragraph{Metrics.}
Pass@1 is the benchmark-native success rate per trajectory, averaged over
all trials; quality differences use a paired bootstrap over tasks
(Appendix~\ref{sec:appendix_quality}). Resource columns are per-trajectory
means over every agent-model request, including Weak drafts, rejected
chunks, verification, repair, and retries. We estimate FLOPs and
HBM-read from each request's token counts and the model architecture
(Appendix~\ref{sec:appendix_flops}, \ref{sec:appendix_hbm}), charging a
prompt prefix once per session under a radix prefix-cache model
\citep{zheng2024sglang} applied identically to all methods
(Appendix~\ref{sec:analysis_ledger}); the HBM interval of the MoE Strong model
reflects its unknown expert routing during prefill. GPU$\cdot$s sums request
duration times serving GPUs, wall time is mean elapsed time per trajectory,
and calls count Weak/Strong generation requests
(Appendix~\ref{sec:appendix_aggregation}).

\begin{table}[H]
\caption{Quality and mean per-trajectory resources of the agent models on
$\tau^2$-Bench, AutomationBench, and VitaBench (Delivery). Compute and HBM-read are estimated from
logged tokens and model architecture, assuming prefix reuse within a task.
HBM-read of a Strong MoE model is a range, because the number of experts a
prefill pass reads is not logged; the dense Weak model has a point estimate.
Calls count Weak and Strong model calls per trajectory. Per benchmark, excluding Weak-only,
the best value in each column except Calls is in bold and the second best underlined.}
\label{tab:main_results}
\centering
\footnotesize
\setlength{\tabcolsep}{2pt}
\begin{tabularx}{\linewidth}{@{}l*{6}{>{\centering\arraybackslash}X}@{}}
\toprule
Method & \shortstack{Pass@1\\(\%) $\uparrow$} &
\shortstack{HBM-read\\(TB) $\downarrow$} &
\shortstack{Wall-time\\(s) $\downarrow$} &
\shortstack{Est. compute\\(PFLOP) $\downarrow$} &
\shortstack{GPU$\cdot$s\\$\downarrow$} &
\shortstack{Calls\\(W/S) $\downarrow$} \\
\midrule
\rowcolor{tablegroupblue}
\multicolumn{7}{@{}l}{\textit{$\tau^2$-Bench (Airline)}} \\
\quad Weak-only (4B) & 80.0 & $83.8$ & 69.7 & 0.18 & 66.0 & $14.1/0.0$ \\
\quad Strong-only (397B) & \textbf{86.0} & $[333.2,341.6]$ & \underline{305.2} & \textbf{0.78} & \underline{4828.4} & $0.0/13.8$ \\
\quad AgentCollab & \underline{85.3} & \underline{$[276.6,289.4]$} & 385.2 & 1.12 & 5629.3 & $9.0/18.5$ \\
\quad Speculative Actions & 72.0 & $[429.5,438.2]$ & 665.6 & \underline{0.89} & 9990.6 & $4.5/17.4$ \\
\quad \textbf{\SpecAgent{}} & \textbf{86.0} & $\mathbf{[147.6,150.1]}$ & \textbf{126.2} & 1.01 & \textbf{860.4} & $15.5/3.1$ \\
\midrule
\rowcolor{tablegroupblue}
\multicolumn{7}{@{}l}{\textit{$\tau^2$-Bench (Retail)}} \\
\quad Weak-only (4B) & 80.4 & $48.4$ & 45.7 & 0.14 & 41.2 & $13.3/0.0$ \\
\quad Strong-only (397B) & \textbf{88.6} & $[192.8,201.8]$ & \underline{162.4} & \textbf{0.58} & \underline{2535.6} & $0.0/13.1$ \\
\quad AgentCollab & 85.7 & \underline{$[171.9,184.2]$} & 223.2 & 0.98 & 3147.8 & $10.3/16.6$ \\
\quad Speculative Actions & 48.5 & $[270.4,280.4]$ & 263.5 & \underline{0.68} & 3944.7 & $5.0/17.1$ \\
\quad \textbf{\SpecAgent{}} & \underline{87.7} & $\mathbf{[107.3,109.7]}$ & \textbf{87.5} & 0.81 & \textbf{789.7} & $13.8/3.1$ \\
\midrule
\rowcolor{tablegroupblue}
\multicolumn{7}{@{}l}{\textit{AutomationBench}} \\
\quad Weak-only (4B) & 6.8 & $94.5$ & 60.1 & 0.29 & 60.0 & $20.4/0.0$ \\
\quad Strong-only (397B) & \textbf{15.2} & \underline{$[324.0,337.3]$} & \underline{307.4} & \textbf{1.10} & \underline{4483.2} & $0.0/17.4$ \\
\quad AgentCollab & 7.3 & $[470.0,510.6]$ & 932.0 & 3.99 & 8603.0 & $15.7/54.8$ \\
\quad \textbf{\SpecAgent{}} & \underline{10.5} & $\mathbf{[264.5,273.6]}$ & \textbf{174.9} & \underline{1.52} & \textbf{1761.0} & $24.4/11.8$ \\
\midrule
\rowcolor{tablegroupblue}
\multicolumn{7}{@{}l}{\textit{VitaBench (Delivery)}} \\
\quad Weak-only (4B) & 34.0 & $77.6$ & 86.3 & 0.21 & 74.2 & $26.0/0.0$ \\
\quad Strong-only (397B) & \underline{48.0} & $[249.6,262.0]$ & 197.3 & \textbf{0.72} & 3018.9 & $0.0/21.9$ \\
\quad AgentCollab & \textbf{50.0} & \underline{$[212.3,220.1]$} & \underline{183.4} & \underline{0.76} & \underline{2329.6} & $19.1/23.9$ \\
\quad \textbf{\SpecAgent{}} & 44.5 & $\mathbf{[154.9,161.1]}$ & \textbf{135.2} & 1.34 & \textbf{814.0} & $25.5/8.2$ \\
\bottomrule
\end{tabularx}
\end{table}
 
\subsection{Main Results}
\label{sec:evaluation_main}

\paragraph{Quality (Q1).}
On $\tau^2$ Airline, \SpecAgent{} reaches the Strong-only Pass@1 (86.0 vs.\
86.0), up from 80.0 for Weak-only. On Retail it comes within 0.9 points of
Strong-only (87.7 vs.\ 88.6, from 80.4 for Weak-only), recovering 89\% of the
Weak--Strong gap. On AutomationBench it recovers 44\% (10.5, between 6.8 and
15.2), and on VitaBench 75\% (44.5, between 34.0 and 48.0). On $\tau^2$, the paired bootstrap
(Appendix~\ref{sec:appendix_quality}) gives a \SpecAgent{}$-$Strong difference
of 0.0 points on Airline (95\% CI $[-5.3,+5.3]$) and $-0.9$ on Retail
($[-4.7,+2.9]$).

\begin{figure}[t]
  \centering
  \includegraphics[width=0.76\linewidth]{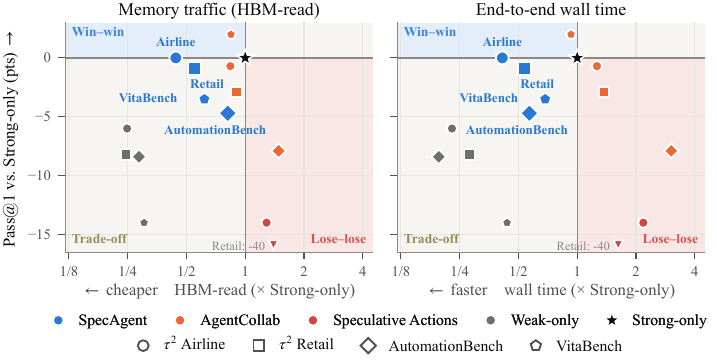}
  \caption{Quality against memory traffic and wall time, relative to each
  domain's Strong-only run (star). Quadrants, drawn from point estimates, mark win--win, lose--lose, and trade-off regions.}
  \label{fig:pareto}
\end{figure}

\paragraph{Resources (Q2): memory traffic and time fall, compute does not.}
Relative to Strong-only (Figure~\ref{fig:pareto}), \SpecAgent{} lowers estimated HBM-read to 43--45\%
on Airline, 53--57\% on Retail, 78--84\% on AutomationBench, and 59--65\% on
VitaBench. Wall time falls to 41\%, 54\%, 57\%, and 69\%. Estimated FLOPs rise by 30\%, 38\%, 38\%, and 87\%. The two axes diverge as Section~\ref{sec:method_cost} anticipates: Strong
calls drop from 13.8 to 3.1 per Airline trajectory, from 13.1 to 3.1 on
Retail, from 17.4 to 11.8 on AutomationBench, and from 21.9 to 8.2 on
VitaBench, so most Strong decoding is replaced by prefill, while Weak drafts
and the Strong review and repair prompts add compute (Section~\ref{sec:analysis}).

\paragraph{Comparison with collaboration methods (Q3).}
Both baselines issue more Strong calls than Strong-only. Speculative Actions
keeps the Strong model acting at every step and adds background Strong
speculation (17.4 vs.\ 13.8 calls on Airline), with more memory traffic and time; in the released implementation it also falls
below Weak-only quality (72.0 on Airline, 48.5 on Retail;
Appendix~\ref{sec:appendix_specactions}). AgentCollab stays near Strong
quality on $\tau^2$ (85.3 and 85.7) with more Strong calls (18.5 vs.\ 13.8 on
Airline) and takes longer. On AutomationBench it
reaches 7.3 with 1.5$\times$ the Strong-only HBM-read and 3.0$\times$ its wall time.
On VitaBench it slightly exceeds Strong-only (50.0 vs.\ 48.0, against 44.5 for
\SpecAgent{}) but uses 1.4$\times$ the HBM-read and wall time of \SpecAgent{}.
\SpecAgent{} matches or exceeds AgentCollab's quality on $\tau^2$ (86.0 vs.\
85.3 and 87.7 vs.\ 85.7) with HBM-read lower by a factor of 1.6--1.9 and
wall time by 2.6--3.1, and it is the only collaborative method whose
HBM-read and wall time are both below Strong-only on every benchmark in
Table~\ref{tab:main_results}.

\subsection{Edge--Cloud Deployment: MCP-Atlas}
\label{sec:evaluation_mcpatlas}

\begin{table}[t]
\caption{Edge--cloud deployment on MCP-Atlas (101 tasks, real MCP servers,
one trajectory per task). The Weak model, Qwen3.5-4B, runs on one local RTX
5090; each Strong model is served by its provider's cloud API. The Weak-only
row is shared by all groups. Resources are per-task means and API cost
is the total over all tasks at list price, excluding the local GPU.}
\label{tab:mcp_atlas}
\centering
\footnotesize
\setlength{\tabcolsep}{2pt}
\begin{tabularx}{\linewidth}{@{}l*{6}{>{\centering\arraybackslash}X}@{}}
\toprule
Method & \shortstack{Pass@1\\(\%) $\uparrow$} &
\shortstack{HBM-read\\(TB) $\downarrow$} &
\shortstack{Strong output\\tokens $\downarrow$} &
\shortstack{Wall-time\\(s) $\downarrow$} &
\shortstack{Est. compute\\(PFLOP) $\downarrow$} &
\shortstack{API cost\\(\$) $\downarrow$} \\
\midrule
Weak-only (4B) & 44.6 & 39.1 & -- & 109 & 0.67 & 0.00 \\
\midrule
\rowcolor{tablegroupblue}
\multicolumn{7}{@{}l}{\textit{Strong: Qwen3.5-397B-A17B}} \\
\quad Strong-only (397B) & 72.3 & $[119.5,128.6]$ & 3.6k & 105 & 1.72 & 1.39 \\
\quad \textbf{\SpecAgent{}} & 59.4 & $[105.7,109.0]$ & 1.8k & 120 & 2.64 & 1.12 \\
\midrule
\rowcolor{tablegroupblue}
\multicolumn{7}{@{}l}{\textit{Strong: GLM-5}} \\
\quad Strong-only (GLM-5) & 76.2 & $[377.6,389.4]$ & 4.6k & 110 & 10.07 & 8.91 \\
\quad \textbf{\SpecAgent{}} & 67.3 & $[217.0,223.1]$ & 2.1k & 108 & 10.50 & 5.54 \\
\midrule
\rowcolor{tablegroupblue}
\multicolumn{7}{@{}l}{\textit{Strong: DeepSeek-V4-Flash-0731}} \\
\quad Strong-only (DeepSeek) & 84.2 & $[150.3,151.7]$ & 10.6k & 151 & 3.38 & 1.37 \\
\quad \textbf{\SpecAgent{}} & 74.3 & $[102.7,103.5]$ & 3.8k & 215 & 2.63 & 0.81 \\
\bottomrule
\end{tabularx}
\end{table}
 MCP-Atlas evaluates multi-step tool use against real Model Context Protocol
servers for search, code execution, databases, and productivity tools
\citep{anthropic2024mcp,bandi2026mcpatlas}; an LLM judge scores each response
by the share of reference claims it covers. We use the 101 public tasks whose
servers run without private API keys (Appendix~\ref{sec:appendix_setup}) and
deploy \SpecAgent{} across the edge and the cloud: Qwen3.5-4B runs locally on
a single RTX 5090, and the Strong model is one of Qwen3.5-397B-A17B, GLM-5
\citep{glm2026glm5}, and DeepSeek-V4-Flash-0731 \citep{deepseek2026v4}, each
behind its provider's API. In this setting the cloud model is billed per token,
so we report its output tokens and API cost alongside quality, and compare
against Weak-only and Strong-only.
Across the three Strong models, \SpecAgent{} recovers 53--75\% of the
Weak--Strong gap while the Strong model decodes only 36--50\% of its
Strong-only output tokens, and API cost falls to 59--81\%
(Table~\ref{tab:mcp_atlas}). For every Strong model the share of the gap
recovered exceeds the share of Strong tokens spent, so each Strong token buys
more quality inside \SpecAgent{} than on its own. Wall
time does not fall here: the local Weak model takes about as long per task as
the hosted Strong models (109 vs.\ 105--151 seconds), so the latency term of
Proposition~\ref{prop:latency} offers no gain, and the benefit appears
instead as lower remote usage and cost.

\section{Analysis and Discussion}
\label{sec:analysis}

We relate the measurements of Section~\ref{sec:evaluation} to the cost model of
Section~\ref{sec:method_cost}, then use it to delimit where \SpecAgent{} pays
off.

\paragraph{Strong decoding explains the saving.}
Proposition~\ref{prop:strong_decode} computes, under
\cref{ass:acceptance}, a quantity that can also be measured directly. Let
$c_v$ and $c_r$ be the verifications and separate repair turns per committed
step, and $\bar v$, $\bar d_r$, $\bar d$ the mean output tokens of a verdict, a
repair, and a Strong-only step. Under \cref{ass:decode_bound} the Strong
model's traffic per committed step, relative to Strong-only, is
\begin{equation}
R=c_v\,\frac{\bar v}{\bar d}+c_r\,\frac{\bar d_r}{\bar d},
\label{eq:strong_accounting}
\end{equation}
of which Proposition~\ref{prop:strong_decode} is the closed form. Multiplying
by the ratio of committed steps per trajectory gives the predicted Strong
share of a trajectory. Table~\ref{tab:strong_accounting} evaluates
\eqref{eq:strong_accounting} from logged token counts. On $\tau^2$ the Strong
model reviews about once every seven committed steps, and its decoded
tokens fall to 17--30\% of Strong-only. The observed Strong share of
HBM-read matches the prediction within 0.01 on every benchmark, slightly
higher because decoded tokens omit the prefill and KV traffic of each call;
generation time matches it on $\tau^2$ and falls below it on AutomationBench
and VitaBench. On AutomationBench, the Weak model's drafts are rejected more often
and each rejection costs a separate repair of about $1.5\bar d$ tokens, so
the $c_r$ term dominates and the saving shrinks, as the $\rho(1-p_u)$ term of
Proposition~\ref{prop:strong_decode} anticipates. VitaBench lies in between:
about 77\% of reviewed boundaries are rejected, but verdicts and repairs
contribute about equally to its predicted share of 0.28.

\begin{table}[b]
\caption{Strong decoding relative to Strong-only: predicted from
\eqref{eq:strong_accounting}, scaled by the ratio of committed steps per
trajectory to that of Strong-only (1.05--1.41), and observed. Time is Strong
generation time relative to that of Strong-only.}
  \label{tab:strong_accounting}
\centering
\small
\setlength{\tabcolsep}{4pt}
\begin{tabular}{@{}lccccccc@{}}
\toprule
& \multicolumn{2}{c}{Verify} & \multicolumn{2}{c}{Repair} & Predicted &
\multicolumn{2}{c}{Observed} \\
\cmidrule(lr){2-3}\cmidrule(lr){4-5}\cmidrule(l){7-8}
& $c_v$ & $\bar v/\bar d$ & $c_r$ & $\bar d_r/\bar d$ & decoded & HBM & time \\
\midrule
Airline & 0.145 & 0.67 & 0.046 & 1.27 & 0.17 & 0.17 & 0.16 \\
Retail & 0.152 & 1.35 & 0.057 & 1.32 & 0.30 & 0.30 & 0.30 \\
AutomationBench & 0.277 & 0.25 & 0.204 & 1.52 & 0.53 & 0.54 & 0.38 \\
VitaBench & 0.216 & 0.53 & 0.100 & 1.36 & 0.28 & 0.29 & 0.24 \\
\bottomrule
\end{tabular}
\end{table}

\begin{figure}[t]
  \centering
  \includegraphics[width=\linewidth]{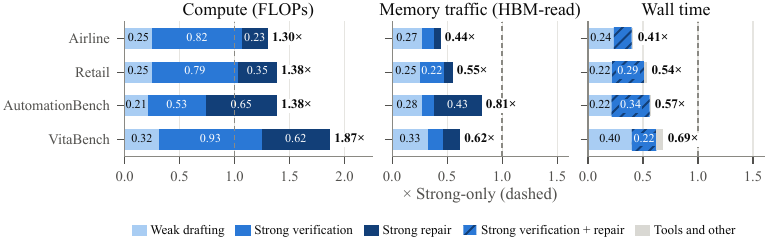}
  \caption{\SpecAgent{}'s cost by phase, relative to Strong-only (dashed).
  Strong verification and repair dominate compute but take a much smaller
  share of memory traffic.}
  \label{fig:phase}
\end{figure}

\paragraph{Why time falls while FLOPs rise.}
As Section~\ref{sec:method_cost} explains, decoding is bound by memory
bandwidth while a verification prefill over a whole chunk is bound by
compute, so re-reading the context in prefill raises FLOPs but sharply reduces
time. Figure~\ref{fig:phase} splits the cost by phase. Strong verification and
repair carry 81--85\% of \SpecAgent{}'s FLOPs but a much smaller share of its
HBM-read (39--66\%); Weak drafting, itself memory-bound decoding, accounts for
only 15--19\% of FLOPs but 34--61\% of HBM-read. Compute nearly
doubles on VitaBench because the verifier and repair prompts each have their
own template and share no prefix with each other or with the actor: the Strong
model prefills about 28.4k uncached tokens per task (12.5k for Strong-only),
1.03~PFLOP that alone exceeds Strong-only's 0.72~PFLOP, while the decoding it
removes is only about 0.21~PFLOP. Wall time decomposes as Proposition~\ref{prop:latency} describes:
Strong generation shrinks to 0.16--0.34 of Strong-only wall time and Weak drafting
adds 0.22--0.24, or 0.40 on VitaBench. On AutomationBench, separate repairs dominate the
Strong model's share, the $\rho(1-p_u)$ term at work.

\paragraph{When verification recovers Strong quality.}
\Cref{ass:acceptance} ties acceptance to how well the Weak model handles a
task, so we split $\tau^2$ tasks by whether Weak-only solves them in all three
trials. Where the Weak model is unreliable, \SpecAgent{} stays close to
Strong-only (70.4 vs.\ 66.7 on Airline, 18 tasks; 74.2 vs.\ 75.8 on Retail, 44
tasks), and where it always succeeds both are near ceiling (94.8 vs.\ 96.9 and
96.2 vs.\ 96.7). These subsets are small, but they suggest that one review and
repair per held action suffices when the Weak model is partly competent; when
it is far weaker, as on AutomationBench, repairs grow and less of the gap is
recovered (44\%); VitaBench falls in between (75\%).

\begin{figure}[t]
  \centering
  \includegraphics[width=\linewidth]{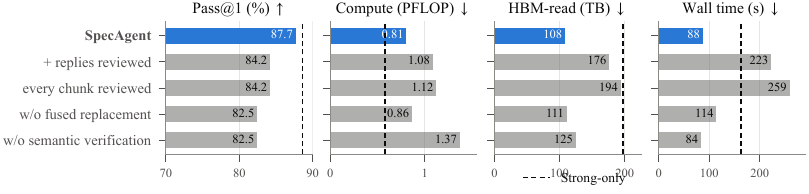}
  \caption{Design ablations on $\tau^2$ Retail, varied one at a time from
  \SpecAgent{}; \SpecAgent{} and Strong-only (dashed) are main-table runs.}
  \label{fig:ablation}
\end{figure}

\paragraph{Design choices.}
Figure~\ref{fig:ablation} varies the three choices of
Section~\ref{sec:method_protocol} on Retail. \emph{Where to review:} also
holding replies, or reviewing every chunk including reads, more than triples
the Strong calls and raises HBM-read by 63--79\% and wall time
$2.5$--$3.0\times$ (past Strong-only), while lowering Pass@1. \emph{How to
repair:} without the fused replacement, every rejection becomes a separate
Strong turn, and Pass@1 drops to 82.5. \emph{What to accept:} a verifier that
accepts only the exact step the Strong model would take itself also lowers
Pass@1 to 82.5 and raises FLOPs by 70\%, because valid Weak steps are
rejected and regenerated. Semantic acceptance thus appears to be what lets verification keep Weak
progress without losing quality.

\paragraph{Across Strong models.}
The MCP-Atlas pairs vary the Strong model with the Weak model fixed
(Table~\ref{tab:mcp_atlas}). \SpecAgent{}'s quality tracks the Strong model's
own (59.4, 67.3, and 74.3 against Strong-only scores of 72.3, 76.2, and
84.2), since the Strong model both judges the drafts and writes the repairs.
Its token share falls as Strong-only trajectories grow longer, as the amortized
verdict of Proposition~\ref{prop:strong_decode} predicts: 50\%, 46\%, and
36\% for 3.6k, 4.6k, and 10.6k Strong-only tokens per task. API cost falls
less (to 59--81\%) because each review resends the context as input, the
prefill side of the trade.

\paragraph{Deployment regimes.}
The two propositions also say when the trade does not pay. The latency term
$\kappa\,n(1-p_u)/(1-p_u^n)$ requires a Weak step to be faster than a Strong
step, which does not hold in the edge--cloud deployment of
Section~\ref{sec:evaluation_mcpatlas}. \Cref{ass:decode_bound} holds while the Strong replica's decode batch stays
below the compute--bandwidth balance point, about 37 sequences on H20; with
larger batches decoding becomes compute bound, and the 30--87\% increase in
FLOPs would begin to cost time.
\section{Conclusion}
\label{sec:conclusion}

\SpecAgent{} turns the Strong model from the author of every agent step into a
reviewer of grounded drafts: speculation runs freely while execution stays
reversible, stops for review only at state-changing actions, and keeps valid
progress rather than only agreement. Across $\tau^2$-Bench, AutomationBench,
VitaBench, and MCP-Atlas, this retains most of Strong-only quality while cutting memory
traffic, wall time, or remote cost, because Strong decoding is replaced by
prefill. The gain grows with the Weak progress the Strong model accepts, so
stronger Weak drafters \citep{xu2026envfactory} and better-calibrated verifiers translate directly
into further memory and latency savings without any change to the protocol.
 
\clearpage

\bibliographystyle{abbrvnat}

\clearpage
\appendix
\section{Experimental Setup}
\label{sec:appendix_setup}

\paragraph{Benchmarks.}
$\tau^2$-Bench Airline (50 tasks) and Retail (114 tasks) are run with three
independent trajectories per task. The simulated user is
DeepSeek-V4-Flash-0731 \citep{deepseek2026v4}, served locally with four-way tensor parallelism and
shared by all methods. AutomationBench uses its 600 public tasks with one
trajectory per task; its simple warm-up domain was used for development and
is excluded. VitaBench uses the 100 tasks of its Delivery domain with two
trajectories per task; its user simulator and rubric judge
are also DeepSeek-V4-Flash-0731.

MCP-Atlas uses 101 of its public tasks whose servers run without private API
keys, excluding six on which a run of any method timed out, and
runs one trajectory per task. The subset is somewhat easier than the full
benchmark (4.5 versus 4.9 ground-truth tool calls and 3.4 versus 3.9 graded
claims per task), so its absolute scores are not comparable to the public
leaderboard; all methods are compared on the same tasks. Responses are graded
by Gemini 3.1 Pro \citep{google2026gemini31pro}, and a task counts as solved
when the response covers at least 75\% of its graded claims.

\paragraph{Models and serving.}
On $\tau^2$, AutomationBench, and VitaBench, the Weak model, Qwen3.5-4B, is served with
SGLang on a single H20 GPU, and the Strong model, Qwen3.5-397B-A17B, runs as a
separate deployment with 16-way tensor parallelism on H20 GPUs. Agents use
the servers' default sampling, with no output-length cap (except 8{,}192
tokens for the Weak model on VitaBench) and at most 100
steps per trajectory. Thinking is enabled whenever a model generates actions,
including Strong repair, and disabled for verification. Every method runs
eight tasks concurrently against the same deployments; on Retail, the Weak
requests of a trajectory are pinned to one replica. On MCP-Atlas, the Weak
model runs on one local RTX 5090 GPU, and each Strong model is called through
its provider's API; verification again runs without thinking.

\paragraph{Baselines.}
Weak-only and Strong-only run the same agent loop as \SpecAgent{} with a
single model. AgentCollab \citep{gao2026agentcollab} follows the published
algorithm: the Strong model acts for the first $K_L=2$ steps, after which the
Weak model acts and each action is followed by the paper's TRUE/FALSE progress
check once its observation arrives. A FALSE check hands control to the Strong
model for $\min(B_{\max},B_0+k\,l)$ steps, with $B_0=2$, $k=2$, $B_{\max}=6$,
and $l$ the number of consecutive failed checks. Because $\tau^2$ agents have
no think tool, the progress check runs as a separate call.
Speculative Actions is described in Appendix~\ref{sec:appendix_specactions}. On MCP-Atlas, where the Strong models are served through third-party APIs, we compare only against the single-model baselines.

\section{Metric Definitions}
\label{sec:appendix}

This section defines every quantity in Table~\ref{tab:main_results}. All
resource columns are means over the official trajectories of a method
(Airline $n=150$, Retail $n=342$, AutomationBench $n=600$, VitaBench $n=200$) and cover the
agent-model requests only; user simulation, judging, and tool serving are
excluded.

\subsection{Task Quality}
\label{sec:appendix_quality}

Let $y_{ir}\in\{0,1\}$ be the benchmark-native success of trial $r$ of task
$i$, with $N$ tasks and $R$ trials per task ($R=3$ on $\tau^2$, $R=2$ on
VitaBench, and $R=1$ on AutomationBench). We report
\begin{equation}
\mathrm{Pass@1}=\frac{1}{NR}\sum_{i=1}^{N}\sum_{r=1}^{R}y_{ir},\qquad
\mathrm{pass@}R=\frac{1}{N}\sum_{i}\max_{r}y_{ir},\qquad
\mathrm{pass}^{R}=\frac{1}{N}\sum_{i}\min_{r}y_{ir}.
\label{eq:pass}
\end{equation}
Failed or incomplete trials stay in the denominator. For the paired
difference $\Delta=\mathrm{Pass@1}^{\SpecAgent}-\mathrm{Pass@1}^{\mathrm{Strong}}$
we draw $N$ tasks with replacement, keep all $R$ trials of each drawn task for
both methods, recompute $\Delta$, and report the 2.5th and 97.5th percentiles
over $10{,}000$ resamples.

\subsection{Request Shape and Prefix Reuse}
\label{sec:analysis_ledger}

Each request $q$ has $I_q$ input and $O_q$ output tokens, of which a prefix of
$C_q$ input tokens is reused from earlier computation. The request then
prefills $P_q$ new tokens, runs $D_q$ decode forwards, and computes $L_q$
output logits:
\begin{equation}
P_q=I_q-C_q,\qquad
D_q=\begin{cases}O_q-1 & P_q>0\\ O_q & P_q=0\end{cases},\qquad
L_q=O_q,
\label{eq:request_shape}
\end{equation}
since the prefill pass already emits the first output token. Serving caches
measure reuse only partially and are affected by eviction, so we assign $C_q$
with a model of radix prefix caching \citep{zheng2024sglang}, applied
identically to every method. Requests are grouped into sessions that share
task, trial, model, role, and serving replica; reuse never crosses sessions.
Within a session, requests are processed in order with the rules of
Table~\ref{tab:radix_rules}.

\begin{table}[h]
\caption{Prefix-reuse rules. $M$, $B^{v}$, and $B^{c}$ are running maxima
kept per session and updated after each request.}
\label{tab:radix_rules}
\centering
\small
\begin{tabular}{@{}lll@{}}
\toprule
Request family & Newly processed tokens $P_q$ & Rationale \\
\midrule
Weak draft, Strong act & $\max(0,\,I_q-M)$ & append-only transcript \\
Strong verification & $c_q+\max(0,\,I_q-c_q-B^{v})$ & candidate $c_q$ is new text \\
Strong repair & $300+\max(0,\,I_q-300-B^{c})$ & rejection note is new text \\
\bottomrule
\end{tabular}
\end{table}

For verification, the candidate length $c_q$ is the Weak output of the same
round (the method's median when unavailable); the remaining backbone of the
prompt can be reused from earlier verifications. A repair prompt adds a
rejection note, for which we assume 300 tokens because note text is not
logged. Verification and repair prompts are not assumed to share a prefix
with each other.

\subsection{Estimated FLOPs}
\label{sec:appendix_flops}

We count a multiply--add as two FLOPs, so a matrix product of shapes
$m\times k$ and $k\times n$ costs $2mnk$. For $T$ token positions whose
full-attention queries see $\Pi$ causal query--key pairs in total, one pass
through the decoder stack costs
\begin{equation}
F_{\mathrm{stack}}(T,\Pi)=\sum_{\ell\in\mathcal L}
\Bigl[F_{\mathrm{mlp}}(T)+
\mathbf{1}[\ell\in\mathcal L_{\mathrm{full}}]\bigl(F_{\mathrm{proj}}(T)+4n_hd\,\Pi\bigr)+
\mathbf{1}[\ell\in\mathcal L_{\mathrm{lin}}]F_{\mathrm{lin}}(T)\Bigr],
\label{eq:flops_stack}
\end{equation}
where $4n_hd\,\Pi$ covers the score and value products of $n_h$ heads of
dimension $d$. The components are
\begin{align}
F_{\mathrm{mlp}}(T)&=\begin{cases}
6Thm & \text{dense SwiGLU}\\
2ThE+6Thm_s+6k\,Thm_e & \text{MoE: router, shared, and $k$ routed experts}
\end{cases}\label{eq:flops_mlp}\\
F_{\mathrm{proj}}(T)&=2Th\,(2n_hd+2n_{kv}d)+2T\,n_hd\,h,\label{eq:flops_proj}\\
F_{\mathrm{lin}}(T)&=2Th\,(2d_K+2d_V+2n_v)+2T\,(2d_K+d_V)\kappa+2Thd_V
+10\,T\,n_vd_kd_v,\label{eq:flops_lin}
\end{align}
where $h$ is the hidden size, $m$ the dense MLP width, $E$, $k$, $m_e$, and
$m_s$ the number of experts, active experts, and routed and shared expert
widths, $n_{kv}$ the number of KV heads, $n_k$ and $n_v$ the Gated DeltaNet key
and value heads with dimensions $d_k$ and $d_v$, $d_K=n_kd_k$, $d_V=n_vd_v$, and
$\kappa$ the convolution width (Table~\ref{tab:arch}). The query projection is
doubled by the output gate, and the last term of
Equation~\ref{eq:flops_lin} is a token-linear proxy for the five Gated DeltaNet
state operations (decay, state-key read, delta, rank-one update, and query
readout). A request then costs
\begin{equation}
F_q=F_{\mathrm{stack}}\bigl(P_q,\;P_qC_q+\tfrac{P_q(P_q+1)}{2}\bigr)
+F_{\mathrm{stack}}\bigl(D_q,\;D_qI_q+\tfrac{D_q(D_q+1)}{2}\bigr)
+2L_qVh,
\label{eq:flops_request}
\end{equation}
for prefill, decode, and the output head. Only the $k$ active experts enter
the MoE term, and FLOPs are not multiplied by the number of serving GPUs.
RMSNorm, elementwise activations, softmax, rotary embeddings, multi-token
prediction, and inter-GPU communication are not counted.

\begin{table}[h]
\caption{Architecture parameters used by the estimators (text backbone).}
\label{tab:arch}
\centering
\small
\begin{tabular}{@{}lcc@{}}
\toprule
& Qwen3.5-4B (Weak) & Qwen3.5-397B-A17B (Strong) \\
\midrule
Hidden size $h$ / vocabulary $V$ & 2560 / 248{,}320 & 4096 / 248{,}320 \\
Layers: full attention / Gated DeltaNet & 8 / 24 & 15 / 45 \\
Query / KV heads, head dim $d$ & 16 / 4, 256 & 32 / 2, 256 \\
DeltaNet value heads $n_v$, $d_k{=}d_v$ & 32, 128 & 64, 128 \\
MLP & dense, $m=9216$ & MoE, $E=512$, $k=10$ \\
Expert / shared width $m_e$ / $m_s$ & -- & 1024 / 1024 \\
Weights / KV / recurrent state & BF16 / BF16 / FP32 & BF16 / BF16 / FP32 \\
\bottomrule
\end{tabular}
\end{table}

The two other Strong models of MCP-Atlas use different attention, so their
estimators replace the attention terms above while keeping the same MLP,
output-head, and weight-read accounting (Table~\ref{tab:arch_api}). GLM-5
\citep{glm2026glm5} uses multi-head latent attention with DSA sparse
attention: a lightning indexer scores every causal query--key pair, and
attention reads only the top-2048 selected keys. DeepSeek-V4-Flash-0731
\citep{deepseek2026v4} combines a 128-token sliding window with compressed
keys and values, at a ratio of 4 or 128 that alternates by layer after two window-only layers, and an
indexer selects the top 512 compressed positions on ratio-4 layers; we also
count its low-rank projections and hyper-connection mixing. Architecture
parameters follow each model's published configuration, multi-token
prediction layers are not modeled, and HBM-read uses the stored precision of
each tensor.

\begin{table}[h]
\caption{Architecture parameters of the MCP-Atlas Strong models GLM-5 and
DeepSeek-V4-Flash-0731 used by the estimators.}
\label{tab:arch_api}
\centering
\small
\begin{tabular}{@{}lcc@{}}
\toprule
& GLM-5 & DeepSeek-V4-Flash-0731 \\
\midrule
Hidden size $h$ / vocabulary $V$ & 6144 / 154{,}880 & 4096 / 129{,}280 \\
Layers & 78 (first 3 dense) & 43 \\
Attention & MLA + DSA & sliding window + compressed \\
Query / KV heads & 64 / 64 & 64 / 1 \\
Head dim (q/k, v) & 256 (192 + 64 rope), 256 & 512 (64 rope) \\
Low-rank projection dims & query 2048, KV 512 & query 1024, output 1024 (8 groups) \\
Indexer heads $\times$ dim, top-$k$ & $32\times128$, 2048 & $64\times128$, 512 \\
MLP & MoE, $E=256$, $k=8$ & MoE, $E=256$, $k=6$ \\
Expert / shared width $m_e$ / $m_s$ & 2048 / 2048 & 2048 / 2048 \\
Dense MLP width $m$ & 12{,}288 & -- \\
Weights / experts / KV & BF16 / BF16 / BF16 & FP8 / FP4 / FP8 \\
\bottomrule
\end{tabular}
\end{table}

\subsection{Estimated HBM-Read}
\label{sec:appendix_hbm}

HBM-read estimates the bytes a request reads from GPU memory when executed
alone (batch size one):
\begin{equation}
B_q=B^{\mathrm{w}}_q+B^{\mathrm{kv}}_q+B^{\mathrm{state}}_q+B^{\mathrm{act}}_q.
\label{eq:hbm_request}
\end{equation}
Weights are read once for the prefill pass and once per decode forward. With
$W(e)$ the weight bytes of the stack when $e$ routed experts per layer are
touched,
\begin{equation}
B^{\mathrm{w}}_q=\mathbf{1}[P_q>0]\,W(U_q)+D_q\,W(k),\qquad
k\le U_q\le\min(E,\,kP_q),
\label{eq:hbm_weights}
\end{equation}
where $U_q$ is the unknown number of distinct experts a layer touches during
prefill; its two bounds give the reported interval. For the dense model,
$W$ does not depend on $e$ and the interval collapses. The remaining terms are
\begin{align}
B^{\mathrm{kv}}_q&=\bigl(D_qI_q+\tfrac{D_q(D_q+1)}{2}\bigr)\cdot 2n_{kv}d\cdot 2\,\mathrm{B}\cdot|\mathcal L_{\mathrm{full}}|,\\
B^{\mathrm{state}}_q&=(P_q+D_q)\cdot n_vd_kd_v\cdot 4\,\mathrm{B}\cdot|\mathcal L_{\mathrm{lin}}|,\\
B^{\mathrm{act}}_q&=(P_q+D_q)\cdot h\cdot 2\,\mathrm{B}\cdot|\mathcal L|,
\end{align}
for the attention cache read during decoding, the recurrent state, and a
token-linear activation proxy. HBM-read excludes writes and inter-GPU
communication and is not multiplied by the number of serving GPUs. It is an
analytical estimate, not a hardware counter measurement. As a spot check,
Nsight Compute DRAM counters for the Weak model on one H20 GPU at batch size
one agree with the estimated weight and KV reads to within 5\% on a held-out
request, and within 10\% across short request shapes; we did not profile the
Strong deployment.

\subsection{Aggregation, GPU Time, Wall Time, and Calls}
\label{sec:appendix_aggregation}

For a method with official trajectories $i=1,\ldots,n$ and successful
generation requests $\mathcal Q_i$, the table reports
\begin{equation}
\overline F=\frac{1}{n}\sum_{i=1}^{n}\sum_{q\in\mathcal Q_i}F_q,\qquad
\overline B=\frac{1}{n}\sum_{i}\sum_{q\in\mathcal Q_i}B_q,\qquad
\overline G=\frac{1}{n}\sum_{i}\sum_{q\in\mathcal Q_i}t_qN_q,
\label{eq:resource_ledger}
\end{equation}
in units of $10^{15}$ FLOPs, $10^{12}$ bytes, and GPU-seconds. Here $t_q$ is
the observed request duration and $N_q$ the GPUs serving it: 16 for the
Strong model (tensor parallelism 16) and one for each Weak request.
$\overline G$ charges every request its full duration on the GPUs that
serve it, as if it ran alone. It is applied identically to all methods, but
because concurrent requests share a deployment and overlap in time, it is an
upper bound on actual occupancy; we use it for relative comparison.
\phantomsection\label{eq:gpu_seconds}
HBM intervals sum the per-request
bounds and are rounded outward.

Wall time is the mean elapsed time per trajectory, from task start to agent
completion on $\tau^2$ and VitaBench, including tool calls and user-simulator turns, and
from the first model call to scoring end on AutomationBench. Calls are the
mean number of successful generation requests per trajectory, split by the
Weak and Strong model; transport retries of one logical request count once.

\subsection{Supplementary Results}

Table~\ref{tab:tau2_solo_quality_n3} reports pass@3 and pass$^3$ for the
$\tau^2$ single-model baselines over the same three trials per task.
\begin{table}[H]
\caption{$\tau^2$-Bench single-model baselines. Pass@1 is mean binary reward
per attempt; pass@3 requires at least one successful attempt per task, and
pass$^3$ requires all three. Scores are percentages. Airline contains 50
tasks and Retail 114 tasks, each with three attempts.}
\label{tab:tau2_solo_quality_n3}
\centering
\small
\begin{tabular}{llccc}
\toprule
Domain & Method & Pass@1 (\%) $\uparrow$ & Pass@3 (\%) $\uparrow$ & Pass$^3$ (\%) $\uparrow$ \\
\midrule
Airline & Weak-only (4B) & 80.0 & 96.0 & 64.0 \\
        & Strong-only (397B) & 86.0 & 96.0 & 72.0 \\
\midrule
Retail  & Weak-only (4B) & 80.4 & 95.6 & 61.4 \\
        & Strong-only (397B) & 88.6 & 96.5 & 76.3 \\
\bottomrule
\end{tabular}
\end{table}
 
\section{Proofs for Section~\ref{sec:method_cost}}
\label{sec:appendix_proofs}

Fix a task $u$ and write $p=p_u$. Consider one cycle: a chunk of length $n$
is drafted, verified, and either committed in full or cut at the first
rejected step and repaired. Let $S$ be the number of steps the cycle commits.
Under \cref{ass:acceptance}, the cycle commits at least $k\le n$ steps exactly
when the first $k-1$ drafted steps are accepted: if $j=n$ all $n$ steps are
committed, and if $j<n$ the $j$ accepted steps and the repair are. Hence
$\Pr(S\ge k)=p^{k-1}$ for $1\le k\le n$ and
\begin{equation}
\mathrm{E}[S]=\sum_{k=1}^{n}p^{k-1}=\frac{1-p^n}{1-p}.
\label{eq:expected_commit}
\end{equation}
A cycle needs a repair when $j<n$, with probability $1-p^n$.

\paragraph{Proof of Proposition~\ref{prop:strong_decode}.}
By \cref{ass:decode_bound}, the Strong cost of a cycle, in weight passes, is
$v$ for the verdict plus $d$ for a separate repair turn, which occurs with
probability $\rho(1-p^n)$. Its expectation is $v+\rho(1-p^n)d$. Within the
task, cycles are independent and identically distributed, so by the
renewal--reward theorem the Strong cost per committed step is this
expectation divided by $\mathrm{E}[S]$. Dividing by the cost $d$ of a
Strong-only step gives
\begin{equation*}
R_u=\frac{v+\rho(1-p^n)d}{d}\cdot\frac{1-p}{1-p^n}
=\frac{1-p}{1-p^n}\cdot\frac{v}{d}+\rho(1-p).
\end{equation*}
Across tasks, \SpecAgent{} spends $T_uR_ud$ weight passes on task $u$ and
Strong-only spends $T_ud$, so the ratio of totals is
$\mathrm{E}_u[T_uR_u]/\mathrm{E}_u[T_u]$.

\paragraph{Proof of Proposition~\ref{prop:latency}.}
The cycle's Strong time is proportional to its weight passes, as above. The
Weak model drafts all $n$ steps before the verdict, taking time $n\kappa$ in
units of a Strong-only step. Calls within a cycle run sequentially, so the
cycle's expected time is $n\kappa+v/d+\rho(1-p^n)$ in the same units, and
dividing by $\mathrm{E}[S]$ yields $L_u=R_u+\kappa\,n(1-p)/(1-p^n)$. The
ratio over tasks follows as for $R$.
\hfill\qedsymbol

\section{Benchmark Adapters}
\label{sec:appendix_adapters}

All benchmarks run the protocol of Algorithm~\ref{alg:specagent}; an adapter
$E$ binds only its benchmark-specific parts. Two bindings are shared. A
fused replacement is accepted only if it matches the tool's argument schema
and executes in the environment ($E.\textsc{Valid}$); otherwise the Strong
model takes one repair turn. The verifier sees committed actions and
observations but not the Weak model's reasoning. Table~\ref{tab:adapters}
lists the bindings that differ.

\begin{table}[h]
\caption{Adapter bindings that differ across benchmarks.}
\label{tab:adapters}
\centering
\footnotesize
\setlength{\tabcolsep}{4pt}
\begin{tabular}{@{}lllll@{}}
\toprule
 & $\tau^2$-Bench & AutomationBench & VitaBench & MCP-Atlas \\
\midrule
Held ($E.\textsc{StateChanging}$) & writes, transfer & writes, final message & writes & writes, final answer \\
Chunk size $K$ & 6 & 8 & 6 & 6 \\
Verifier context & full history & full history & tool results cut & old tool results cut \\
\bottomrule
\end{tabular}
\end{table}

\paragraph{Choices.}
Tools are classified as reads or writes from each benchmark's documentation
and their observed effect on the environment, or, on VitaBench, from its own
read/write tool labels. Rejecting a read is a logical
de-commitment: the read is removed from the agent's history and from what the
Strong model reviews, but an external service may still keep traces of it,
such as access logs. The guarantee is therefore that no action classified as
state-changing executes before review, not that every effect of speculation
is physically rolled back. On $\tau^2$-Bench and VitaBench, replies to the
simulated user cannot be undone but are not held: they leave the database
unchanged, the user's next turn returns as an observation, and on $\tau^2$
holding them more than triples Strong calls without improving quality
(Figure~\ref{fig:ablation}).
On AutomationBench and MCP-Atlas the final message is the graded answer, so
it is held like a write, and on MCP-Atlas it is reviewed against the evidence
gathered by earlier reads. MCP-Atlas and VitaBench tool outputs can be long, so
tool results are truncated in the verifier's view (older ones only on
MCP-Atlas). $K$ bounds how many reads
accumulate before an unreviewed commit.

\subsection{Prompts}
\label{sec:appendix_prompts}

On $\tau^2$, Weak drafting and the single-model baselines use the benchmark's
default agent prompt with the domain policy. The Strong verifier and the
separate repair turn use the prompts below, where \texttt{\{domain\}} and
\texttt{\{domain\_policy\}} are filled with the domain name and the
benchmark's policy text. The other benchmarks use the same
structure with their own policies and tools. All prompts are released with
the code.

\paragraph{Verifier.}
{\footnotesize
\begin{verbatim}
You are the Strong trajectory reviewer for a Weak-led customer-service agent
in the "{domain}" domain. Assess whether the draft is a justified continuation,
retain its longest valid prefix, and redirect it when necessary.

1. INPUT AND EVIDENCE

- Input: task, confirmed_history, proposed_trajectory, candidate_length K,
  invoked_tool_contracts, and the domain policy below.
- Candidate steps are indexed 1..K. pending=true means the step has not
  happened (a staged mutation or an unsent reply). A step without pending
  has already executed; its observation is real.
- Ground judgments in original observations. Assistant hypotheses and
  review notes are not factual evidence.

Domain policy:
{domain_policy}

2. REVIEW THE DIRECTION

Evaluate the chunk against the customer's request, the policy, and visible
history. Notice ignored contradictions, mutations whose preconditions are
unmet, replies that claim work that was not done, and unproductive repeats.

Customer-service replies:
- Asking for a user id, reservation id, date, or name that is not yet in
  the task, customer messages, or tool observations is a justified
  continuation. Do not reject it for failing to finish the request.
- A reply that promises a cancellation, refund, booking change, waiver,
  or other state change is a commit. Reject it unless that mutation is
  already in confirmed_history or this chunk with a successful observation.
- When the policy requires explicit customer confirmation before acting,
  a draft that mutates without that answer is unjustified. "The customer
  implied yes" is not confirmation.

Policy:
- A mutation or a denial is justified only by a sentence in the domain
  policy above. If you cannot point to that sentence, do not invent a rule.
- If two policy-legal outcomes remain, do not reject the draft for picking
  one unless the conversation plus policy uniquely selects the other.

Executed reads (pending=false lookups):
- The observation is already real. Rejecting the read cannot undo it; it
  only discards later steps. Do not zero- or partial-accept a reads-only
  prefix because a later mutation would be better, because the agent is
  "still searching", or because the matching reservation has not turned
  up yet. Searching the customer's existing reservations for the one they
  described is justified.
- Reject an executed read only when the tool or required arguments are
  invalid: unknown tool, missing required argument, or an identifier that
  never appeared in the task, customer messages, or prior observations.

Pending mutations (pending=true tool calls):
- Judge the call as if it will run. Wrong user id, reservation id, payment
  id, amount, cabin, or date is a reject even if a later undo exists.

3. RETAIN THE LONGEST VALID PREFIX

Set accepted_steps to the largest j for which steps 1..j form a justified
continuation, stopping before the first unjustified action.
- Judge each action using evidence available before it.
- j applies only to the candidate chunk. j=0 discards that chunk, not
  confirmed_history.
- Stop at unknown tools, invalid required arguments, policy-illegal
  mutations, or a reply that ends the turn with required work missing.

4. EXPLAIN THE REJECTION WHEN j < K

Write reason as a concise, evidence-based decision brief: what is wrong,
why continuation stops at this prefix, and what the next step should
establish. The next actor receives confirmed_history, steps 1..j, and
reason, without the discarded suffix. Do not present facts found only in
the discarded suffix as retained evidence.
If you also emit replacement_action, reason only diagnoses the discarded
suffix; it must not assign extra work to a second Strong actor.

5. PROPOSE ONE TOOL ACTION OR HAND OFF

- If you can specify a useful next tool call from the retained context,
  you MUST return it as replacement_action: {"tool": "...", "arguments": {...}}.
  Returning null spends a second Strong call. Use null only when the next
  step is a customer reply, or you cannot fill schema-valid arguments from
  the retained prefix.
- This field is tool-only: one call, a supplied tool, complete schema-valid
  arguments. Mutating tools are allowed when the policy's preconditions
  are already satisfied by the retained prefix.
- Do not return null "to be safe" when a lookup tool already has its
  arguments in the retained observations.

6. OUTPUT

Return exactly one JSON object with exactly these three keys:
{"reason": <string or null>, "accepted_steps": <integer 0..K>,
 "replacement_action": <one tool action or null>}
- Full acceptance (j == K): reason and replacement_action must both be null.
- Partial acceptance or rejection (j < K): reason must contain the diagnosis;
  replacement_action follows section 5.
- Do not emit rejection_code, corrector, or any other key.
\end{verbatim}
}

\paragraph{Repair.}
{\footnotesize
\begin{verbatim}
You are the Strong customer-service agent continuing from a reviewed
trajectory. Reassess the conversation and produce exactly one useful next
step. You are invoked only when replacement_action was null or failed the
schema guard; do not re-verify the retained prefix.

1. INPUT AND EVIDENCE

- You receive the original task and tools, the retained dialogue, and
  review.reason. The retained dialogue is confirmed history plus the
  accepted prefix, with actions and original observations.
- Treat the review as a diagnosis to examine, not as new factual evidence.
  Check its material claims against retained observations; when they
  conflict, follow the observations.

2. CHOOSE ONE STEP

- Reply when the required work for this customer turn is done, or when
  the policy requires asking the customer something you do not have.
- Prefer a missing lookup or a required confirmation question over a
  speculative mutation.
- Otherwise issue one tool call targeting the blocking evidence gap or
  the policy-legal mutation whose preconditions are already met.
- Follow the supplied schema. Do not invent identifiers.

3. OUTPUT

Produce exactly one assistant step: either one tool call or one reply.
Do not mention the reviewer, the discarded draft, or this protocol.
\end{verbatim}
}

\section{Speculative Actions on \texorpdfstring{$\tau^2$}{tau2}-Bench}
\label{sec:appendix_specactions}

Speculative Actions \citep{ye2026specactions} targets latency rather than
Strong-model work. An authoritative Actor produces every committed step,
while a faster Speculator predicts upcoming environment steps and executes
them in advance. In the authors' $\tau$-bench retail setting, the Speculator
predicts the user's next query and the API calls needed to answer it; the
paper describes validating these guesses once the real query arrives and
restricting speculative side effects to idempotent, reversible, or sandboxed
actions, and it reports API-prediction accuracy rather than task success.

We ported the authors' released $\tau$-bench agent to $\tau^2$-Bench with our
model pair. The Strong model acts at every step. Whenever it replies to the
user, the Weak model role-plays the next user message in the background, and
a Strong decider continues from that guess for up to five tool steps; the
resulting tool calls and observations are added to the context on the next
user turn. As in the released code, these background tool calls execute in
the live environment without a read-only filter, and their results are added
without checking the guess against the real user message.

This configuration reached 72.0\% Pass@1 on Airline and 48.5\% on Retail,
below Weak-only (80.0\% and 80.4\%), while issuing 17.4 and 17.1 Strong calls
per trajectory, more than Strong-only (13.8 and 13.1). The extra Strong calls
come from the background decider, which runs on top of the Strong actor.
Because background calls may change environment state and add observations on
the basis of a guessed user message, the released implementation is not
lossless in this setting, which plausibly explains part of the quality drop;
we did not diagnose it further. These numbers therefore characterize the
released implementation rather than the guarded design described in the
paper.
 
\end{document}